\documentclass{article}
\usepackage{amsmath,epsfig}
\usepackage[preprint]{spconfa4}
\usepackage{amsmath,amssymb,amsfonts}
\usepackage{graphicx,subfigure}
\usepackage{textcomp}
\usepackage{bbm}
\usepackage{caption}
\usepackage{booktabs}
\usepackage{color}
\usepackage{array}
\usepackage{multirow,graphicx}
\usepackage{subfigure}
\usepackage{amssymb}
\usepackage{makecell}
\usepackage{xspace}
\usepackage{xcolor}
\usepackage{bbding}
\usepackage{pifont}
\usepackage{wasysym}
\usepackage{amsmath,graphicx,multirow,algorithm,algorithmic,bm}
\usepackage[colorlinks,linkcolor=black]{hyperref}
\usepackage{wasysym}

\usepackage{cite}

\copyrightnotice{978-1-6654-3864-3/21/\$31.00~\copyright 2021 IEEE}

\let\OLDthebibliography\thebibliography
\renewcommand\thebibliography[1]{
	\OLDthebibliography{#1}
	\setlength{\parskip}{0pt}
	\setlength{\itemsep}{0pt plus 0.3ex}
}

\begin{document}\sloppy

% Example definitions.
% --------------------
\def\x{{\mathbf x}}
\def\L{{\cal L}}

% Title.
% ------
\title{Unsupervised Remoting Sensing Super-Resolution Via Migration Image Prior}
%
% Single address.
% ---------------
\name{Jiaming Wang $^{1}$, Zhenfeng Shao $^{1}$, Tao Lu $^{2} \textsuperscript{*}$, Xiao Huang $^{3}$, Ruiqian Zhang $^{1}$, and Yu Wang $^{2}$  \thanks{This work was supported in part by the National key R \& D plan on strategic international scientific and technological innovation cooperation special project under Grant 2016YFE0202300, the National Natural Science Foundation of China under Grants 61671332, 41771452, 51708426, 41890820,  62072350, and 41771454, the Natural Science Fund of Hubei Province in China under Grant 2018CFA007, the Independent Research Projects of Wuhan University under Grant 2042018kf0250.}}

%Address and e-mail should NOT be added in the submission paper. They should be present only in the camera ready paper. 
\address{$^{1}$ Wuhan University, $^{2}$ Wuhan Institute of Technology, $^{3}$ University of Arkansas\\
\{wjmecho, shaozhenfeng, zhangruiqian\}@whu.edu.cn, lutxyl@gmail.com\\ xh010@uark.edu, wangyu949374585@gmail.com}

\maketitle
\vspace{-0.4cm}

\begin{abstract}
Recently, satellites with high temporal resolution have fostered wide attention in various practical applications. Due to limitations of bandwidth and hardware cost, however, the spatial resolution of such satellites is considerably low, largely limiting their potentials in scenarios that require spatially explicit information. To improve image resolution, numerous approaches based on training low-high resolution pairs have been proposed to address the super-resolution (SR) task. Despite their success, however, low/high spatial resolution pairs are usually difficult to obtain in satellites with a high temporal resolution, making such approaches in SR impractical to use. In this paper, we proposed a new unsupervised learning framework, called ``MIP'', which achieves SR tasks without low/high resolution image pairs. First, random noise maps are fed into a designed generative adversarial network (GAN) for reconstruction. Then, the proposed method converts the reference image to latent space as the migration image prior. Finally, we update the input noise via an implicit method, and further transfer the texture and structured information from the reference image. Extensive experimental results on the Draper dataset show that MIP achieves significant improvements over state-of-the-art methods both quantitatively and qualitatively. The proposed MIP is open-sourced at \url{https://github.com/jiaming-wang/MIP}.
\end{abstract}
\vspace{-0.4cm}
\section{Introduction}
Recently, remote sensing satellites, which are especially appropriate for uninterrupted observing targets, have drawn widespread concerns in various practical applications. Continuously monitoring moving targets by high temporal resolution satellites, can expand the application range than satellites with a static image, such as, the Jilin-1 and Zhuhai-1 OVS-1 A/B video satellites.  It is common knowledge that the spatial resolution and spectral resolution are always a pair of contradictory for the optical remote sensor. Additionally, due to bandwidth and hardware cost limitations, the spatial resolution of high temporal resolution satellite images is decreased that cannot meet the demand of high precision applications. Therefore, improving the spatial resolution of satellite images with large compression ratios, has become an urgent issue in remote sensing applications.

Super-resolution (SR) aims to reconstruct the high spatial resolution (HSR) image from observed low spatial resolution (LSR) images \cite{park2003super}, which breaks the limitations of the imaging system for the best cost/benefit ratio. In real-world remote sensing scenarios, the SR problems often have the following properties: 1) HSR and high temporal resolution (HSR-HTR) datasets are unavailable, 2) HSR and low temporal resolution images (HSR-LTR), which enjoy the same image content with LSR high temporal resolutions imageries (LSR-HTR), are easy to obtain as the reference, 3) there are obvious differences between the imaging environments of the LSR-HTR and HSR-LTR images, which makes it difficult to transfer texture directly.

\begin{figure}[t]
	\centering
	\includegraphics[height=4cm]{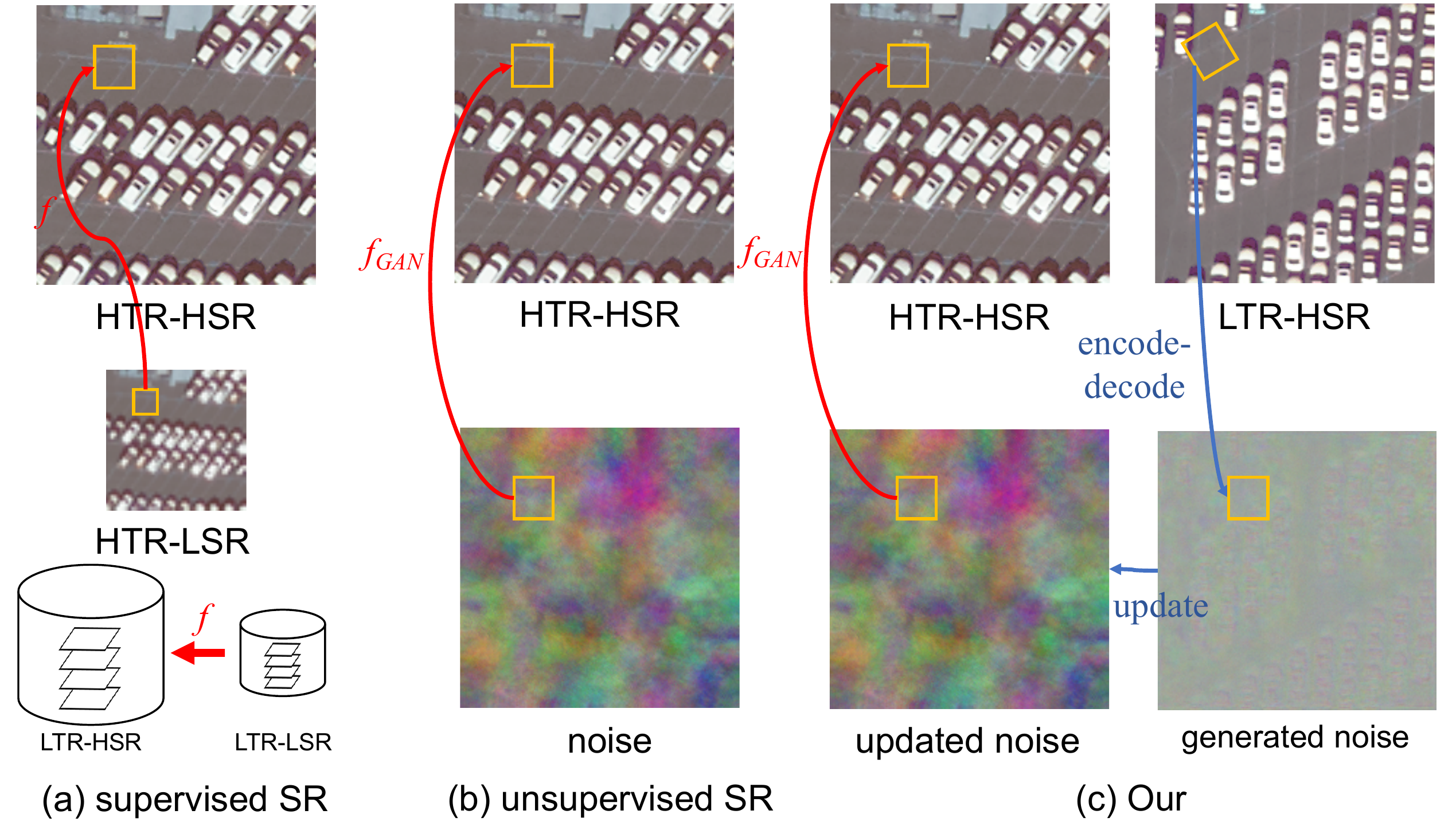}
	\caption{Comparison with supervised and unsupervised SR frameworks. Left column: Input the image pairs and learn the mapping between LTR-HSR and LTR-LSR. Middle column: Generate HTR-HSR images from random noise. Our method performs geometric transformation and texture transfer from the LTR-HSR, and estimates HTR-HSR.}
	\label{com}
\end{figure}

Existing SR methods tend to generate HSR images/patches from the prior information provided by datasets. The development of machine learning promotes the progress of SR. Traditional supervised deep-learning-based SR methods obtain excellent performance by designing a network to extract deep features, \emph{i.e.}, enhanced deep residual networks \cite{lim2017enhanced} and residual dense network \cite{zhang2018residual}. The rationale of these algorithms can be summarized as follows: a deep model learns the mapping between the corrupted LSR images and HSR ones by a convolutional neural network (CNN), and the LSR images are degraded from their original versions. Although these methods are intended for obtaining deep features from image prior information, they have the main disadvantage: they require HSR training examples, which are limited by economic and technical reasons in remote sensing \cite{haut2018a}.

Under the above circumstances, unsupervised image SR has received more attention. To exploit the prior structure, several unsupervised encoder-decoder-based approaches have been developed. Haut \emph{et al.} \cite{haut2018a} proposed an unsupervised deep network for generating remote sensing images from random noise, and improved the resolution of remote sensing imagery, which proved that a generator network is sufficient to capture low-level image statistics prior to any learning in image restoration task \cite{lempitsky2018deep}. However, the performance of unsupervised SR methods is limited, and it is difficult to recover the high-frequency information of the image from the existing image prior.

Inspired by the development of reference-based image super-resolution (RefSR) methods, such as \cite{ wang2017the, zheng2018crossnet}, we intend to investigate unsupervised reference-based strategies to overcome this obstacle. However, the reference dataset \cite{wang2016event} is taken within a short time interval (3 hours) from the social network, which means a stable angle and environment for imaging. The imaging range of HTR satellite images is wide and the time interval between the LSR-HTR and the HSR-LTR images is larger. In this study, we propose a novel unified framework for the high-temporal image SR method. The major benefit of the used encoder-decoder model with spatial transformer networks \cite{jaderberg2015spatial} in the proposed method is that we aggressively learn transformer parameters, which can align the LSR image and the reference image from the huge difference brought by satellite perspective. At the same time, the image is converted to high-dimensional space, and the attributes of the reference image are extracted. Then, the feature map of the reference image is transformed into the latent space which is leveraged as a migration image prior for input noise updating. We conduct experiments to demonstrate the superiority of the proposed framework. Experimental results demonstrate that the proposed method obtains more realistic images, and outperforms state-of-the-art methods.

Fig. \ref{com} schematically illustrates the important steps in our algorithm, and compares it with traditional SR frameworks. The objective of existing supervised single image SR methods (Fig. \ref{com} (a)), is to learn the optimal mapping function between the LTR-HSR and LTR-LSR images from the dataset. Fig. \ref{com} (b) demonstrates the processes of generating imagery from a noise map in unsupervised methods. Different from traditional models, the proposed method firstly introduces the reference image in remote sensing SR task, and converts the reference image into the latent space as the migration image prior. 

The main innovative contributions of this paper are two folds:
\begin{enumerate}
	\item To the best of our knowledge, the proposed method is the first developed approach for unsupervised high-temporal remote sensing images SR, which employs a similar HSR-LTR image as the reference to generate similar HSR texture information.
	\item We propose a novel model which transforms the reference into the latent space as the migration image prior. It provides state-of-the-art results than existing unsupervised SR methods.
\end{enumerate}

\begin{figure}[h]
	\centering
	\includegraphics[width=9cm]{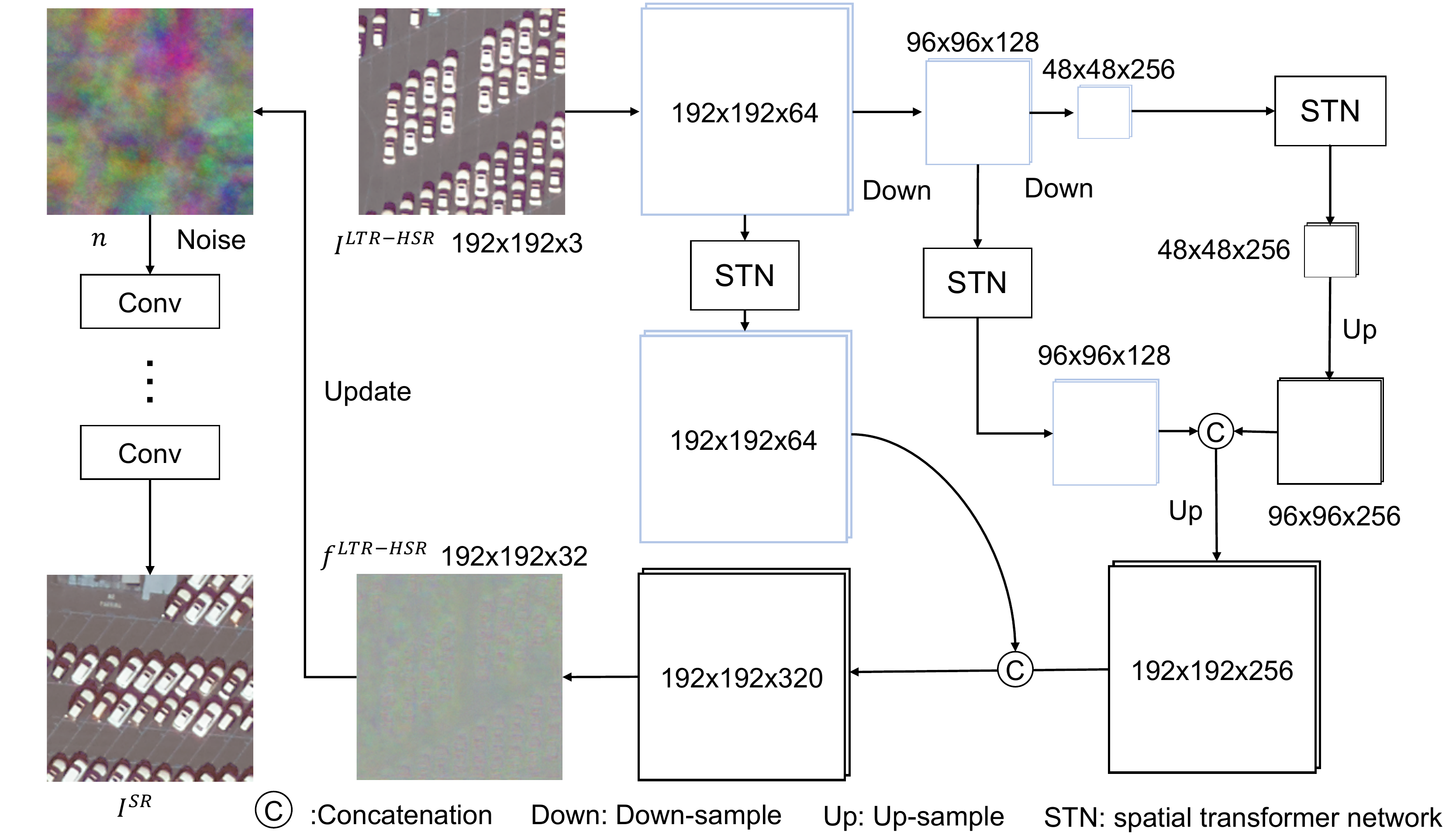}
	\caption{Illustration of the proposed method.}
	\label{network}
\end{figure}
\vspace{-0.4cm}
%The remainder of this paper is organized as follows. Section \ref{s2} provides a brief review of related works. In Section \ref{s3}, we describe the proposed method in detail. Section \ref{s4} reports the experimental results. Finally, we conclude this paper in Section \ref{s5}.

%Traditional universal image/video SR algorithms often down-sample the original HR image and generate LR images. Most existing video SR methods adopt a time- \cite{sajjadi2018frame, tao2017detail} or sapce-based \cite{jo2018deep} ways for utilizing temporal information.

%In real-world satellite scenarios, the super-resolution (SR) task have the following properties: 1) high-resolution (HR) video data is unavailable, however, HR images data is available, 2) degradation method is unknown.

\vspace{-0.4cm}
\section{Related Work}\label{s2}

\subsection{Satellite Image Super Resolution}
Satellite image SR is a very challenging problem. The early study of SR mainly focuses on construct shallow-learning-based models. Merino and Nunez \cite{merino2007super} proposed a linear reconstruction method to combine different LR images. Yang \emph{et al.} \cite{yang2011novel} firstly introduced the dictionary learning into the satellite SR task. However, these shallow networks are difficult to recover the high-frequency information with non-linear scenarios, and the performance is limited.

Recently, Luo \emph{et al.} \cite{luo2017video} proposed a mirroring reflection method to avoid the loss of images' border information. At the same time, considering the lack of high-resolution video data, Gaofen-2 images and Jilin-1 video imageries are used as training and testing samples respectively in \cite{luo2017video}. Moreover, Lu \emph{et al.} \cite{lu2019satellite} considered a framework for fusing multi-scale information in the residual domain, which effectively enhanced the high-frequency information. Jiang \emph{et al.} \cite{jiang2019edge} proposed a generative adversarial network (GAN) based edge-enhancement method that can generate clean and sharp details. Haut \emph{et al.} \cite{haut2018a} firstly proposed an unsupervised hourglass model to super-resolved LSR remote sensing images from random noise. However, it is difficult to recover the high-frequency information of the image from the existing image prior.

\vspace{-0.4cm}

\subsection{Reference-based Image Super-Resolution}

Different from single image super-resolution (SISR) methods, RefSR algorithms provide more accurate and realistic details, which are transferred from the reference image (the reference is similar to LSR one in content, but with different focal lengths and shot perspectives).

Considering the incomplete coupling of the LSR and reference image, some algorithms \cite{wang2017the, zheng2018crossnet} achieved great performance when they are tightly aligned. This means they only swap the information in the image level. In view of this, Zhang \emph{et al.} \cite{zhang2019image} proposed a deep model and adopted local texture matching for long-distance dependency. Most recently, Yang \emph{et al.} \cite{yang2020learning} introduced a more accurate way to search and transfer relevant textures from Ref to LSR images. SSEN \cite{shim2020robust} aligned the Ref and LSR images in the feature domain to capture similarity-aware. In general, these deep methods achieve better results than SISR methods.

However, the improvements of RefSR methods \cite{maeda2020unpaired,yuan2018unsupervised} rely on lots of training images. At the same time, due to different synthetic bands, some satellite images used in this paper (Draper) show different visual characteristics. It is difficult to transform the reference image feature into the input image.

\vspace{-0.4cm}
\section{Our Method}\label{s3}

\subsection{Problem Formulation}\label{s30}

Focusing on the primary goal of SR, to recover the high-frequency information from LSR-HTR images $ \bm{I}^{LSR-HTR} \in {\mathfrak{R} ^{C \times H \times W}}$ and obtain an HSR-HTR version image $ \bm{I}^{HSR-HTR} \in {\mathfrak{R} ^{C \times t \cdot H \times t\cdot W}}$, the conventional formulation of SR methods is ${\bm{I}^{LSR-HTR}} = \bm{D}{\bm{I}^{HSR-HTR}}$, where $\bm{D}$ denotes the down-sampling matrix, and $t$ is the factor. Here, we assume that the paired HSR-LSR high-temporal training data are unavailable, which makes them with simulated paired data impractical. Nevertheless, we can obtain a set of HSR low-temporal reference images that can be used for unsupervised training. Rather than minimizing the error between the SR images and the ground truth in the supervised method, the proposed method is based on the reference image to the texture and content information. Therefore, the key issue of the proposed method is to explore a unified framework to fuse the information at different times.
%Traditionally, multi-frame based supervised training methods require data that is rigid or nonrigid alignment \cite{capel2003computer}. 

The pipeline of MIP is summarized as Fig. \ref{network}. We denote $ \bm{I}^{Ref} \in {\mathfrak{R} ^{C \times t\cdot H \times t\cdot W}}$ the corresponding HSR reference image  $ \bm{I}^{HSR-HTR}$. The random noise maps $\bm{n}_{init}$ is $C' \times t\cdot H \times t\cdot W$. The proposed method mainly consists of three parts: the generative network, the reference feature extraction network and the migration image prior model. First, we learn a mapping from noise maps to an HR image. Second, we adopt an encoder-decoder model to code and transform the reference image. In the end, we map the coded feature maps of the reference image into the latent space, and update the random noise $\bm{n}_{init}$. Details are given in the following.

\vspace{-0.4cm}
\subsection{Image Generation}\label{s31}

Different from GAN-based image generation tasks, SR requires the result as real, not just a high-quality image. If we directly apply image generation \cite{goodfellow2014generative} or GAN-based SR \cite{ledig2017photo} models, we need to up-sample the input, which will also cause the obvious checkerboard phenomenon in the unsupervised framework.

\begin{figure}[h]
	\centering
	\includegraphics[height=2cm]{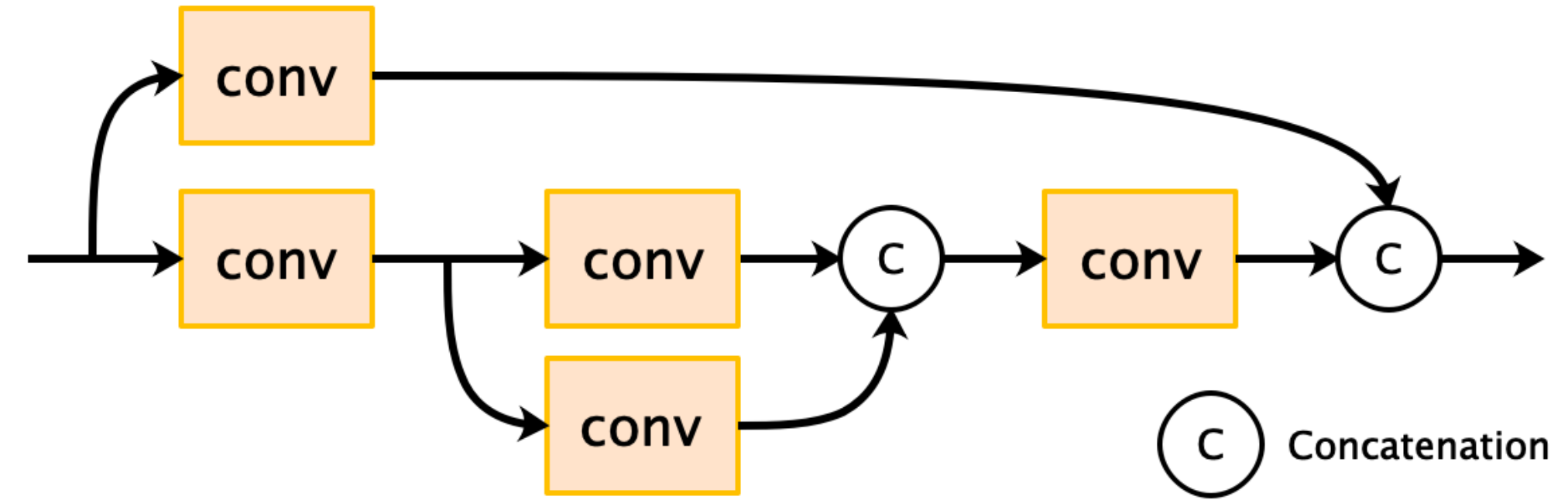}
	\caption{Illustration of the skip model in the generative network.}
	\label{gnetwork}
\end{figure}

Given an input HSR-sized noise maps to generate an image. Hence, it can be formulated as,
\begin{equation}
\bm{I}^{SR}= H(\bm{n}_{init}),
\end{equation}
where $\bm{n}_{init}$ is the noise maps, and $H(.)$ denotes the function of the SR network in the proposed method. $\bm{I}^{SR}$ refers to the output of the SR network. In this paper, we adopt stacked skip models for reconstructing, as shown in Fig. \ref{gnetwork}.  For the proposed skip model, the first extract the shallow feature maps and concatenate it with deep-level features. Another advantage of the skip model is that it can reduce the cost of calculation than the densely connected convolutional network \cite{huang2017densely}.

\vspace{-0.1cm}

\vspace{-0.4cm}

\subsection{Reference Feature Extraction}\label{s32}

Traditional reference-based supervised approaches all try to design deep networks, and align the reference and SR images in the feature domain  \cite{yang2020learning, zhang2019image}. Therefore, the performance of these algorithms is highly dependent on the pixel by pixel supervision. Considering the universal rigid transformation in satellite images, which as taken at different times, the images will have a great difference in shooting angle. In this paper, we advocate an encoder-decoder-based model to exploit the prior of the reference image.

Different from previous, which transform the local features in the reference image into the SR image, we employ spatial transformer networks (STN) \cite{jaderberg2015spatial} to improve the invariance of the affine transformation of a CNN network. It is trained with learnable localisation and grid, as the affine transformation matrix. Then, the image sampling function is used to sample feature maps, and merge with them into a spatial transformer. In this work, we leverage the STN block for the transcoding process.

The spatial transformer network is defined as follows:
\begin{equation}
\left[ {\begin{array}{*{20}{c}}
	{{\bm{x}^{output}}}\\
	{{\bm{y}^{output}}}
	\end{array}} \right] = \left[ {\begin{array}{*{20}{c}}
	{{\theta _{11}}}&{{\theta _{12}}}&{{\theta _{13}}}\\
	{{\theta _{21}}}&{{\theta _{22}}}&{{\theta _{23}}}
	\end{array}} \right]\left[ {\begin{array}{*{20}{c}}
	{{\bm{x}^{input}}}\\
	{{\bm{y}^{input}}}\\
	1
	\end{array}} \right],
\end{equation}
where $({\bm{x}^{input}}, {\bm{y}^{input}})$ is the coordinates in the input feature maps, $({\bm{x}^{output}}, {\bm{y}^{output}})$ is the coordinates in the output maps, and $\theta$ denotes the 2D transformation parameters. MIP consists of several STN blocks as shown in Fig. \ref{network}. MIP gradually aligns reference features in each scales. These blocks facilitates the method to transform feature maps for semantic aligning.

\subsection{Image Prior Migration}\label{s33}

%The unsupervised method allows us to learn the distribution of the data and the correlations between the random noise maps $\bm{n}$ and $\bm{I}^{SR}$. 
We then investigate how to explore the prior in a reference image. Considering the weak supervision in this framework than traditional approaches, it is difficult to transform texture from the reference with a great difference. In the previous work \cite{radford2015unsupervised}, the authors introduce the influence of latent space on the generated results, which laid the solid foundation for fine image generation. Both target and attribute in the results can be mapped to noise vector. InfoGAN \cite{chen2016infogan} decomposes the input noise vector into random noise and the latent code, which can target the structured features. With fixed noise, InfoGAN learns interpretable representations by manipulating latent code.

In this paper, we now propose a method for implicit updating: we convert the feature maps of the reference image into latent space, which carries structural information and code for texture generation. And then the code is used to generate the same semantic targets and attributes. It can be formulated as,
%Generative-based unsupervised SR networks all try to obtain touching details from a random noise matrix.

%In this paper, we advocate a heuristic updating strategy, which can be formulated as,

\begin{equation}
\bm{n} \leftarrow \Psi ( {\bm{n}_{init}, \bm{f}^{Ref}}),
\end{equation}
where $\bm{f}^{Ref}$ is the feature maps of the reference image. In particular, the input noise is updated:
\begin{equation}
\bm{f}(x)=\frac{1}{\sqrt{2 \pi} \operatorname{std}\left(\bm{f}^{Ref}\right)} \exp \left(-\frac{\left(x-\operatorname{mean}\left(\bm{f}^{Ref}\right)\right)^{2}}{2 \operatorname{std}\left(\bm{f}^{Ref}\right)^{2}}\right),
\end{equation}
where $std(.)$ denotes the standard deviation function, and $mean(.)$ is the mean function. $\bm{f}(x)$ is the hidden space matrix generated by the Gaussian function, which conducts the migration image prior from the reference image. Then, the input can be viewed as the combination of the initialized noise and the latent code.

\begin{equation}
\bm{n}_{i+1}=\bm{n}_{init}+\alpha \cdot \bm{f}(x), and \ \bm{n}_{1} = \bm{n}_{init},
\end{equation}
where $i$ is the number of iterations. The updated noise maps will be used for the input of the generation network. As a matter of experience, $\alpha$ is 0.03.

\subsection{Loss Function}\label{s34}

In the current literature, the goal of supervised SR is to generate an HSR image/patch from LSR one, and minimize the error in the HSR space. Considering the lack of ground truth, we downsample the SR image $\bm{I}^{SR}$ by the Lanczos resampling \cite{turkowski1990filters} function as \cite{haut2018a}, and minimize the mean squared error (MSE) between it with the LSR-HTR image ${\bm{I}^{LSR-HTR}}$ in LSR domain. This process can be described as follows:
\begin{equation}
\begin{array}{l}
{\cal L}(\theta ,S) = {\left\| {{\bm{I}^{LSR-HTR}} - down({\bm{I}^{LSR-HTR}})} \right\|_2}\\
\qquad \quad \ \ \, = {\left\| {{\bm{I}^{LSR-HTR}} - {\bm{I}^{LSR-HTR'}}} \right\|_2},
\end{array}
\end{equation}
where $\theta$ denotes the parameters in the proposed method, and $S$ is the training data. $\bm{I}^{LSR-HTR'}$ denotes the LSR-sized version SR image. The Lanczos kernel can be described as follows: 
\begin{equation}
L(\bm{x}) = \frac{{3\sin (\pi \bm{x})\sin (\pi \bm{x}/3)}}{{{\pi ^2}{\bm{x}^2}}},
\end{equation}
where $\bm{x}$ in the input pixel.

\vspace{-0.1cm}
\section{Experiments}\label{s4}

\subsection{Datasets}
The Draper dataset\footnote{https://www.kaggle.com/c/draper-satellite-image-chronology/data} is a publicly available benchmark for remote sensing image ordering in southern California, including 324 scenarios with 5 images in each scenario. The photographs were captured from a plane as a reasonable facsimile for satellite images, which were taken at different times. The HR image size is $3,099 \times 2,329$ pixels. We randomly select two sets of images (five images in a group) from this dataset, and name them ``Day 1'', ``Day 2'',``Day 3'',``Day 4'', and ``Day 5''. It is noteworthy that this is not necessarily a consecutive time. We select 115 LSR version images from ``Day 5'', and corresponding ones in ``Day 4'' as the reference images.

\vspace{-0.3cm}
\begin{figure}[h]
	\centering
	\includegraphics[width=8cm]{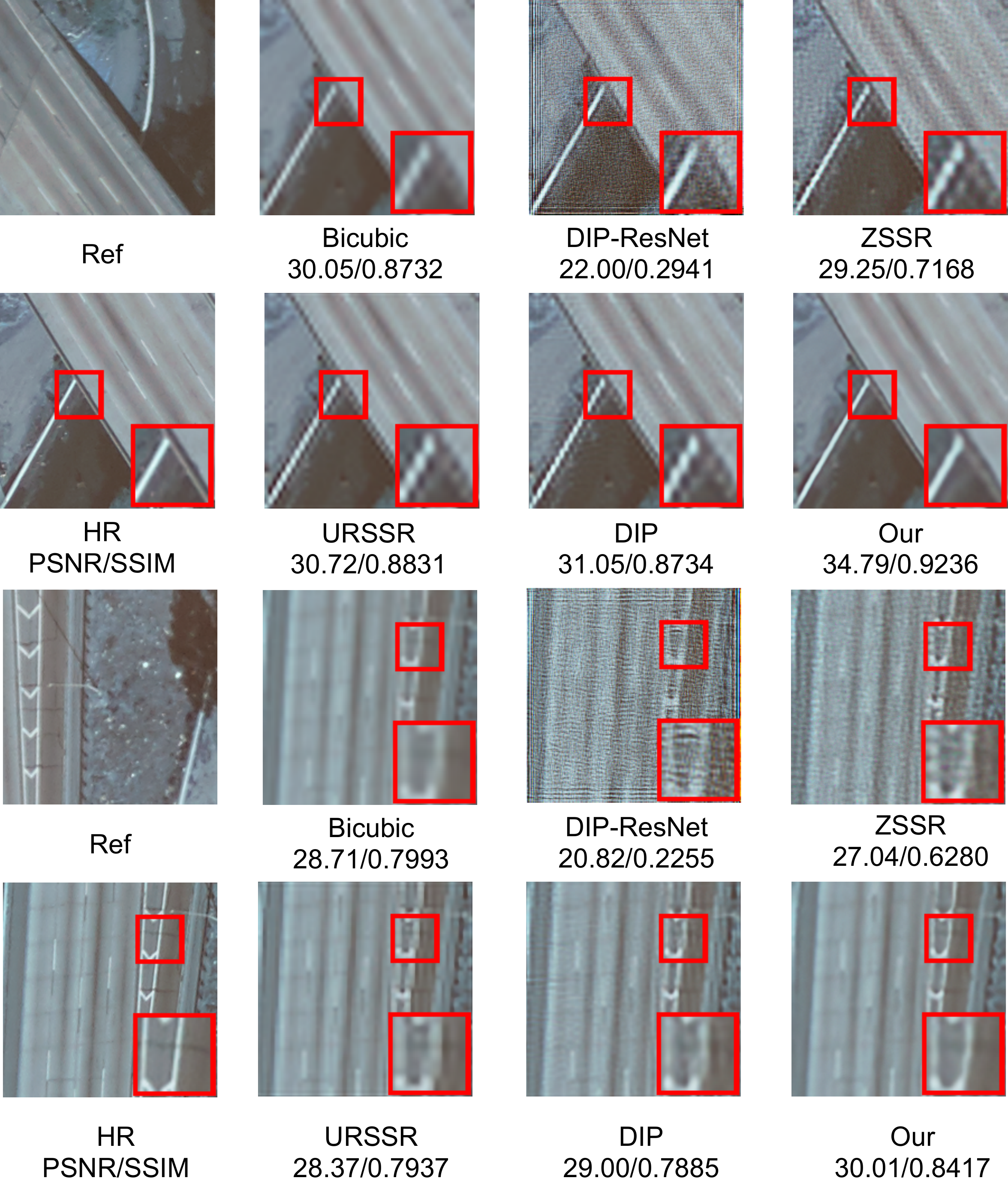}
	\caption{Visual comparison among different SR methods on draper dataset with scale factor $\times$4. We report the PSNR (dB), and SSIM results of the competing methods. The proposed method achieves state-of-the-art performance.}
	\label{result}
\end{figure}
\subsection{Implementation Details}

All the models presented in this paper are trained with Adam optimizer with ${\beta _{\rm{1}}}{\rm{ = 0}}{\rm{.9}}$, ${\beta _{\rm{2}}}{\rm{ = 0}}{\rm{.999}}$, and $\epsilon = 1{\rm{e}} - 8$. Each mini-batch contains one noise map with size $192 \times 192$ and the Ref patches with size $192 \times 192$. We initialize learning rate to $1e-4$. We set the spectral bands of noise $C' = 32$. These experiments run at a desktop with two NVIDIA GTX 2080Ti GPUs and 3.60 GHz Intel Core i7-7820X CPU, 32GB memory. We implement the proposed method using PyTorch 1.1.0 library with Python 3.5.6 under Ubuntu 18.04, CUDA 10.1, and CUDNN 7.5 systems. We train the model over 10000 iterations, until it converges.

\textbf{Evaluation measures.} Four widely used image quality assessment indices are employed to evaluate the performance, including peak signal to noise ratio (PSNR), structural similarity (SSIM), visual information fidelity (VIF) \cite{sheikh2006image}, and erreur relative globale adimensionnelle de synthese (ERGAS).

\subsection{Comparison with Unsupervised Methods}
We compare the results of the proposed method with those of state-of-the-art unsupervised SR methods, DIP \cite{lempitsky2018deep}\footnote{https://github.com/DmitryUlyanov/deep-image-prior}, URSSR \cite{haut2018a}, and ZSSR \cite{ZSSR}\footnote{https://github.com/assafshocher/ZSSR}, among which URSSR \cite{haut2018a} is considered to achieve state-of-the-art performance in remote sensing image SR. DIP \cite{lempitsky2018deep} has achieved the state-of-the-art visual quality, even if compared with the supervised SR algorithm. All experiments are performed with $\times4$. For a fair comparison, all methods are trained with only the input image.

\begin{table}[h]
	\centering
%	\normalsize
	\caption{Average quantitative comparisons of different approaches with scale factor $\times$4.}
	\renewcommand\arraystretch{1.0}
	\begin{tabular}{c|c|c|c|c}
		\hline
		Method   & PSNR $\uparrow$     & SSIM $\uparrow$     & VIF $\uparrow$     & ERGAS $\downarrow$  \\ \hline \hline
		Bicubic & 28.79 & 0.7910 & 0.4018 & 1.6029  \\  \cline{1-5} 	
		DIP-ResNet     & 14.36 & 0.1964 & 0.0812 & 7.9412 \\  \cline{1-5} 	   
		ZSSR   & 28.99 & 0.7487 & 0.3194 & 1.5508 \\  \cline{1-5} 
		URSSR    & 29.29 & 0.8095 & 0.3996 & 1.5215 \\ \cline{1-5} 
		DIP      & 29.63 & 0.8114 & 0.3869 & 1.4530 \\ \cline{1-5} 
		Ours    & \textbf{30.55} &\textbf{0.8388} & \textbf{0.4453} & \textbf{1.3266} \\ \cline{1-5} \hline	
	\end{tabular}
	\label{t1}
\end{table}

Table \ref{t1} shows the average performance of the PSNR, SSIM, VIF, and ERGAS results of competing methods with $\times4$ on the draper dataset. Clearly, the proposed MIP framework outperforms all other competing methods. On average, the PSNR and SSIM values of the proposed MIP framework for upsampling factor $t=4$ are 0.92/1.26 dB and 0.0274/0.0293 higher than the second-best method, respectively.

Several subjective results with upsampling factors $t=4$ are illustrated in Fig. \ref{result}. From the visual reconstruction results, we can see that DIP-ResNet \cite{lempitsky2018deep} and ZSSR \cite{ZSSR} achieve not only shape edge (high-frequency information), but also a lot of noise. DIP \cite{lempitsky2018deep} and URSSR \cite{haut2018a}, which are designed for the unsupervised SR, fail to generate stable and touching detail information. We think this is mainly due to the limitation of the unsupervised SR task. Our method produces sharper edges and finer details than the other methods.
\vspace{-0.1cm}

\section{Conclusion}\label{s5}
\vspace{-0.1cm}

In this paper, we introduce an unsupervised reference-based image SR method termed Migration Image Prior (MIP). In particular, in order to solve the problem of missing high spatial resolution data of high temporal resolution images, we carefully design an end-to-end framework to fully exploit the available high spatial resolution image as the reference. In addition, we adopt a novel way to update input noise, which is used to generate a corresponding high-resolution image. In this way, we encode the reference image into the latent space as the migration image prior, and update noise maps to obtain stable results. Experimental results on the public dataset demonstrate that our method achieves state-of-the-art performance quantitatively and qualitatively.
\vspace{-0.4cm}
%\section*{Acknowledgements}\label{s6}
%\vspace{-0.4cm}
%This work was supported in part by the National key R \& D plan on strategic international scientific and technological innovation cooperation special project under Grant 2016YFE0202300, the National Natural Science Foundation of China under Grants 61671332, 41771452, 51708426, 41890820,  62072350, and 41771454, the Natural Science Fund of Hubei Province in China under Grant 2018CFA007, the Independent Research Projects of Wuhan University under Grant 2042018kf0250.
%\vspace{-0.4cm}

\bibliographystyle{IEEEbib}
\bibliography{icme2021template}

\end{document}